\documentclass[letterpaper, 10 pt, conference]{ieeeconf}  

\usepackage{times}
\usepackage[numbers]{natbib}
\usepackage{multicol}
\usepackage[bookmarks=true]{hyperref}

\usepackage{lipsum}
\usepackage{comment}
\usepackage{balance}
\usepackage{graphicx}
\usepackage[table]{xcolor}
\usepackage{caption}
\usepackage{listings}
\usepackage{xcolor}
\usepackage{upquote} 

\definecolor{codeblue}{RGB}{75,156,211}
\definecolor{codegreen}{RGB}{77,153,0}
\definecolor{codepurple}{RGB}{155,81,224}
\definecolor{codegray}{RGB}{128,128,128}
\definecolor{backcolour}{RGB}{245,245,245} 

\usepackage{xcolor}
\definecolor{bg}{HTML}{f2f2ea}
\usepackage{float} 

\definecolor{bg}{rgb}{0.95, 0.95, 0.92}

\usepackage{seqsplit}

\IEEEoverridecommandlockouts                              

\usepackage{tabularx} 

\title{\Large \bf
APOLLO Blender: A Robotics Library for Visualization and Animation in Blender
}

\author{\authorblockN{Peter Messina and Daniel Rakita}
\authorblockA{Department of Computer Science, 
Yale University}
\authorblockA{\{peter.messina, daniel.rakita\}@yale.edu}
}

\begin{document}

\maketitle
\thispagestyle{empty}
\pagestyle{empty}


\begin{abstract}
High-quality visualizations are an essential part of robotics research, enabling clear communication of results through figures, animations, and demonstration videos. While Blender is a powerful and freely available 3D graphics platform, its steep learning curve and lack of robotics-focused integrations make it difficult and time-consuming for researchers to use effectively. In this work, we introduce a lightweight software library that bridges this gap by providing simple scripting interfaces for common robotics visualization tasks. The library offers three primary capabilities: (1) importing robots and environments directly from standardized descriptions such as URDF; (2) Python-based scripting tools for keyframing robot states and visual attributes; and (3) convenient generation of primitive 3D shapes for schematic figures and animations. Together, these features allow robotics researchers to rapidly create publication-ready images, animations, and explanatory schematics without needing extensive Blender expertise. We demonstrate the library through a series of proof-of-concept examples and conclude with a discussion of current limitations and opportunities for future extensions.
\end{abstract}

\section{Introduction}
\label{sec:introduction}

Clear and compelling visuals are central to the communication of robotics research. Whether in the form of figures for papers, animations for submission videos, or explanatory graphics for presentations, the ability to convey complex concepts visually is critical for both technical audiences and broader outreach. While modern 3D graphics software such as Blender provides powerful tools for creating high-quality renderings and animations, these platforms are rarely designed with robotics in mind. As a result, researchers often face steep barriers when attempting to generate robotics-specific visuals, requiring significant effort to learn software workflows, manually configure robots, and avoid visual inconsistencies or bugs.

Existing solutions within the robotics community only partially address this challenge. Game engines such as Unity or Unreal have been adopted for real-time simulation and interactive demonstrations, but they require extensive development time to achieve high-quality renderings. Dedicated simulators such as Gazebo, CoppeliaSim, or Webots offer robust physics engines and standard robot description support, yet they typically prioritize accurate dynamics over visualization fidelity. Specialized visualization tools, meanwhile, are often restricted to educational use cases or narrow application domains, limiting their flexibility and visual quality. Across this spectrum, there remains a gap between tools optimized for physical accuracy and tools optimized for high-quality static and animated renderings.

In this work, we present \textit{APOLLO Blender}, a lightweight software library that lowers this barrier by providing robotics-focused scripting tools within Blender. Our goal is not to replicate the full functionality of physics simulators or game engines, but rather to make it straightforward for researchers to create static images, animations, and schematic illustrations that are both visually appealing and robotics-aware. The library centers on three key features: importing robots and environments from standardized descriptions, keyframing robot configurations and appearance attributes, and generating common 3D primitives such as lines and cubes to support schematics. Together, these capabilities streamline the visualization workflow, reducing the time and expertise required to produce publication-ready results.

\begin{figure}[t!]
    \centering
    \includegraphics[width=\linewidth]{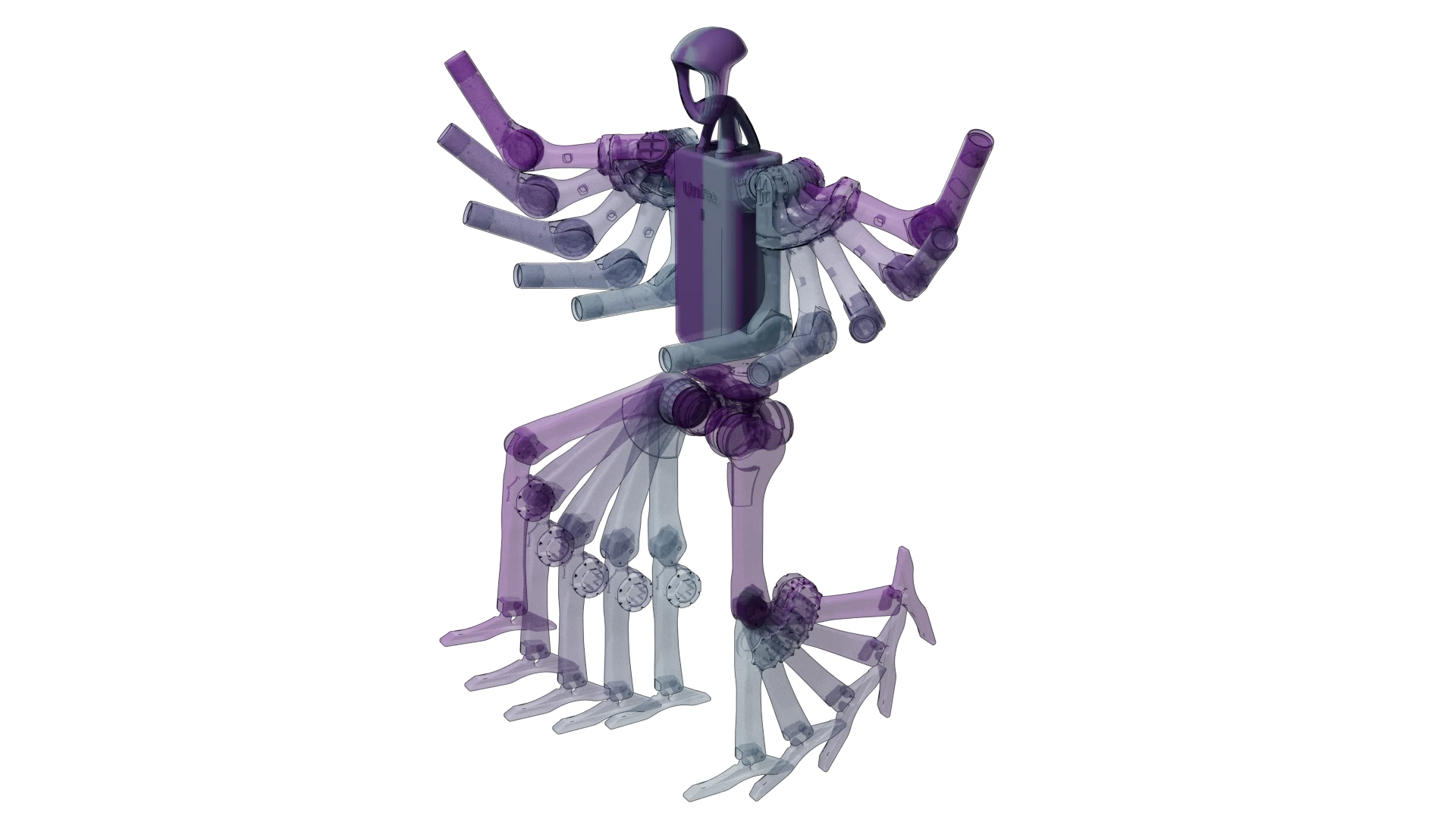}
    \caption{In this paper, we introduce \textit{APOLLO Blender}, a lightweight Python library that enables robotics-aware visualization and animation directly within Blender. The image above shows a humanoid robot executing a complex motion sequence, rendered as a single static frame using multiple transparent, color-faded instances.  By leveraging standard URDF descriptions and concise scripting interfaces, our tool supports rapid generation of publication-quality figures and animations tailored for robotics research, instruction, and outreach.}
    \label{fig:teaser}
    \vspace{-0pt}
\end{figure}

To encourage adoption and extensibility, we are releasing this library as open source. The Python-based tools are provided as pip-installable packages, allowing researchers to integrate them seamlessly into existing Blender workflows with minimal setup. We demonstrate the library’s use through a series of proof-of-concept examples, ranging from robot motion visualizations to schematic figures and geometric approximations. By providing robotics researchers with accessible scripting interfaces for Blender, our system expands the range of tools available for visual communication and aims to accelerate the integration of high-quality graphics into robotics research.

\section{Related Works}
\label{sec:related_works}

In this section, we review prior approaches to robotics visualization across three categories: the use of general-purpose game engines, robotics-focused simulators, and specialized visualization tools. By situating our system within this landscape, we highlight the gap between accurate simulation platforms and high-quality rendering tools that our work aims to address.

\subsection{Game Engines in Robotics}
Game engines are widely used in robotics research for visualization and physics simulation \citep{bartneck2015robot, saivichit2024robotic, pukki2021unreal, li2024photorealistic, lixingjian2024photorealistic, wolf2020evolution, young2020unreal, morse2021gaming}. Unity has been employed to support human–robot interaction research through platforms such as the Robot Engine \citep{bartneck2015robot}, and to illustrate kinematic concepts in interactive environments \citep{mattingly2012robot}. These works highlight Unity’s strengths in real-time animation, interaction design, and a user-friendly interface.

Unreal Engine has also seen growing adoption, particularly in agricultural robotics and autonomous ground vehicle research. Its ability to simulate messy field conditions, dynamic lighting, and lightweight physics at high frame rates has made it suitable for these domains \citep{li2024photorealistic, lixingjian2024photorealistic, young2020unreal}. However, achieving photorealism typically requires substantial modeling effort and the construction of custom datasets, limiting accessibility for general-purpose robotics visualization.

In contrast to these real-time engines, our system focuses specifically on creating high-quality static renderings and animations. Game engines optimize for interactive performance, often at the expense of lighting fidelity, texture resolution, or ease of setup. By leveraging Blender’s offline rendering capabilities, our library prioritizes photorealism and clarity of presentation rather than simulation speed. This distinction makes it particularly well-suited for generating figures and videos intended for scientific communication.

\subsection{Simulators in Robotics}
Robotic simulators such as Gazebo, CoppeliaSim, Webots, and others are designed to replicate complex robot dynamics and interactions \citep{ivaldi2014tools, de2019analysis, farley2022pick, andreiev2024comparative, nogueira2014comparative}. These tools emphasize accurate modeling of physics, contact, and control, often offering multiple physics engines to balance realism and performance. For example, CoppeliaSim has been shown to accurately capture mobile robot motion across varied terrains by allowing users to select among physics engines \citep{farley2022pick}. Similarly, Gazebo has been favored for humanoid simulation due to its open-source ecosystem, active community, and optimized solvers for multi-body contact \citep{ivaldi2014tools}.

While these simulators are indispensable for testing and validation, their visual output is generally secondary to physics fidelity. As a result, rendered images and animations often lack the polish of dedicated graphics tools. Our system complements these simulators by targeting the visualization stage: it imports standard robot descriptions such as URDF, but instead of focusing on accurate physics, it enables researchers to quickly generate clear, visually appealing renderings and animations for communication purposes.

\subsection{Robot Visualization Tools}
Several specialized software packages exist to support robot visualization and education \citep{manseur2004visualization, sadanand2015virtual, gil2015development, miller2005robotic}. For instance, RobotDraw was developed as an interactive platform for teaching manipulator kinematics, allowing students to directly explore virtual robot behavior \citep{manseur2004visualization}. RoboAnalyzer, built on the MATLAB Robotics Toolbox, provides pre-built functions for modeling and analyzing serial-chain robots, lowering the barrier for users without robotics expertise \citep{sadanand2015virtual}. GraspIt! extends these capabilities to robotic grasping, enabling detailed analysis and visualization of hand–object interactions \citep{miller2005robotic}.

These systems demonstrate the value of accessible visualization interfaces, but they are typically limited in scope, tied to specific robotics domains, or focused on educational contexts. Visual fidelity is often modest, and their user communities remain relatively small. By embedding robotics-aware functionality into Blender, our system connects roboticists to a large, established computer graphics ecosystem. This integration provides both ease of use and access to advanced rendering features, expanding the possibilities for producing high-quality, publication-ready visualizations.
\section{Implementation Overview}
\label{sec:implementation_overview}

In this section, we provide an overview of the design and implementation of our system. The goal is to outline the core components, describe how they integrate with Blender, and highlight the scripting interfaces that enable robotics researchers to create high-quality visualizations. We first discuss the overall software structure, then detail the installation and setup process, and finally describe the key scripting functions that form the backbone of the library.

\subsection{Code Structure and Setup}

The library is designed to integrate seamlessly with Blender's existing Python environment. Installation requires only a single step: the package can be directly pip-installed into the Python interpreter bundled with Blender. No additional dependencies or external toolchains are needed, which keeps the setup lightweight and accessible across different platforms.

Once installed, the workflow is straightforward. Blender already includes a built-in Python scripting environment, so users can directly add short scripts inside Blender or save them as external files to be loaded later. A script can, for example, load a robot specification in URDF format, after which the library automatically parses the URDF, extracts both geometric meshes and kinematic structure, and spawns the robot directly into the Blender scene. From this point, the robot can be configured, animated, or modified using simple scripting functions provided by the library. This organization minimizes boilerplate setup and ensures that researchers can move quickly from installation to generating high-quality visualizations.

A central goal of our library is to make Blender’s rendering and animation capabilities directly accessible to robotics researchers through simple Python scripting. To this end, the library provides a suite of functions organized into three categories: (1) robot and environment importing; (2) configuration and keyframing of robot states and visual attributes; and (3) generation of primitive 3D shapes. These functions are designed to minimize boilerplate code, allowing users to quickly move from importing a robot to producing high-quality visualizations.

\subsection{Robot Importing}

Robots and environments can be imported directly into Blender using a Universal Robot Description Directory (URDD)~\cite{klein2025urdd}. A URDD can be constructed automatically from a standard URDF.  This process automatically loads both the geometric meshes that define the robot’s appearance and the kinematic structure that governs its configuration. By relying on URDD, researchers can reuse existing specifications without modification, ensuring compatibility with a wide range of commonly used platforms.

A typical script to spawn a robot requires only a few lines of code. The example below demonstrates how a URDD can be loaded and visualized:

\begin{figure}[H]
    \centering
    \begin{lstlisting}[label=lst:spawn_robot]
from blender_robot_toolbox_py.prelude import *
from blender_robot_toolbox_py.prelude_bpy import *

# Access robot resources
r = ResourcesRootDirectory.new_from_default()  
s = r.get_subdirectory('ur5')             # load robot
c = s.to_chain_numpy()                    # convert urdf
blender_robot = ChainBlender.spawn(c, r)  # Spawn robot\end{lstlisting}
    \vspace{-2pt}
    \caption{Implementation of robot spawning using the toolbox.}
    \label{fig:spawn_robot_code}
    \vspace{-10pt}
\end{figure}

\begin{figure}[t!]
    \centering
    \includegraphics[width=\columnwidth]{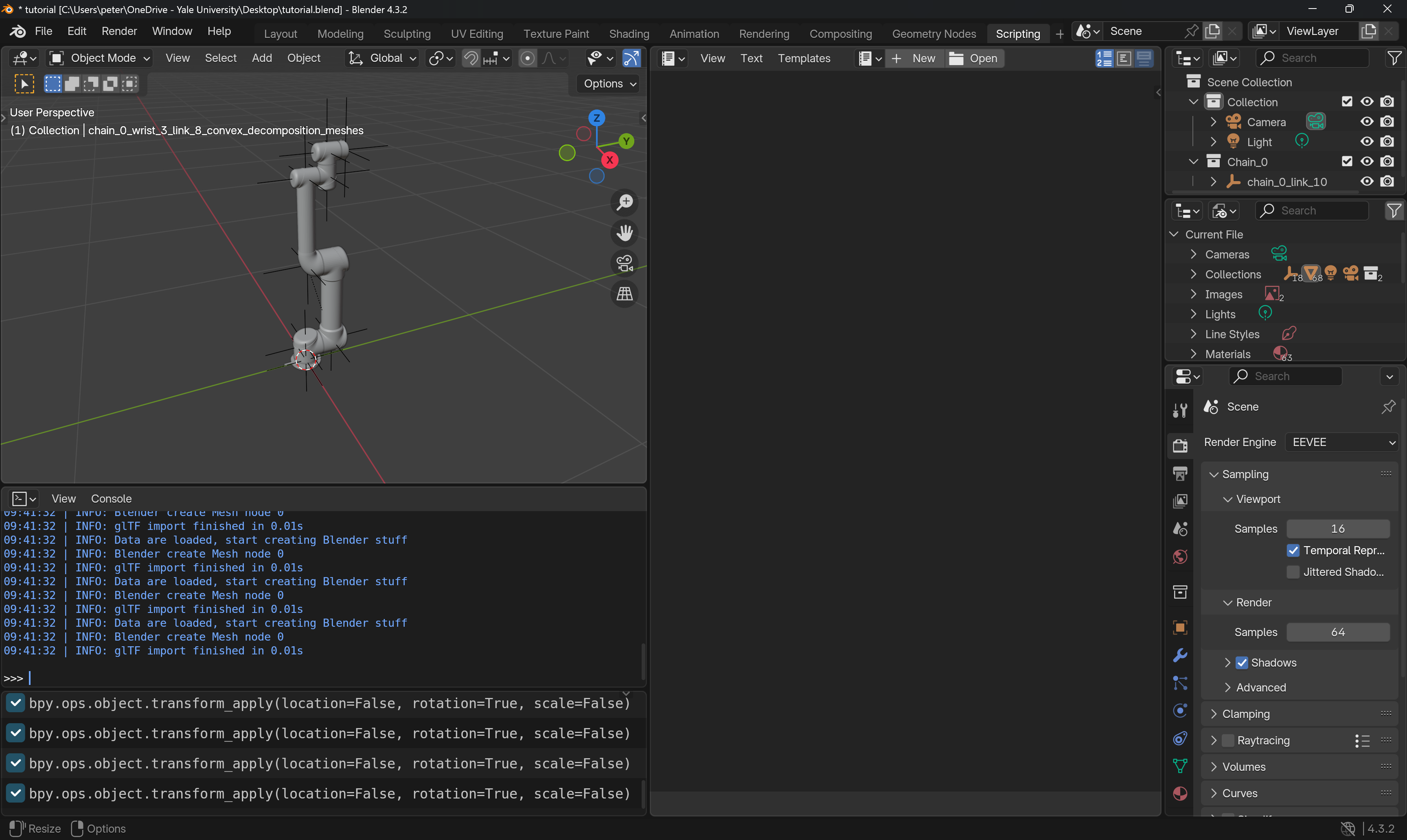}	
    \caption{Example of a UR5 robot spawned into Blender via a short Python script.}
    \label{fig:spawn_robot}
\end{figure}

Here, the function \texttt{new\_from\_default()} points to a standard directory on the user’s computer where robot models are stored. This allows researchers to maintain a consistent and easily accessible location for commonly used URDDs. In this example, the script locates the UR5 model, parses its specification into a kinematic chain, and spawns the robot directly into the Blender scene (Figure~\ref{fig:spawn_robot}).

In addition to custom URDFs, the library also supports pre-defined robot models, allowing users to quickly prototype visualizations without first preparing their own specifications.

\subsection{Robot Configuration}

Once a robot has been imported into Blender, researchers can easily configure it by specifying joint states in the robot’s native joint space. The library provides a simple interface for updating joint positions, enabling rapid prototyping of different poses and motions.

\begin{figure}[H]
    \vspace{-10pt}
    \centering
    \begin{lstlisting}[
        language=Python,
        label={lst:robot_config}
    ]
blender_robot.set_state(state_array)\end{lstlisting}
    \vspace{-2pt}
    \caption{Setting the joint configuration for the robot instance.}
    \label{fig:robot_config_code}
    \vspace{-10pt}
\end{figure}

The \texttt{set\_state()} function accepts an array of joint values (in radians) whose length corresponds to the number of degrees of freedom (DOF) of the robot. For convenience, the library includes utility commands such as \texttt{c.dof\_module}, which return the DOF directly and help users construct valid state arrays.

\begin{figure}[t!]
    \centering
    \includegraphics[width=\columnwidth]{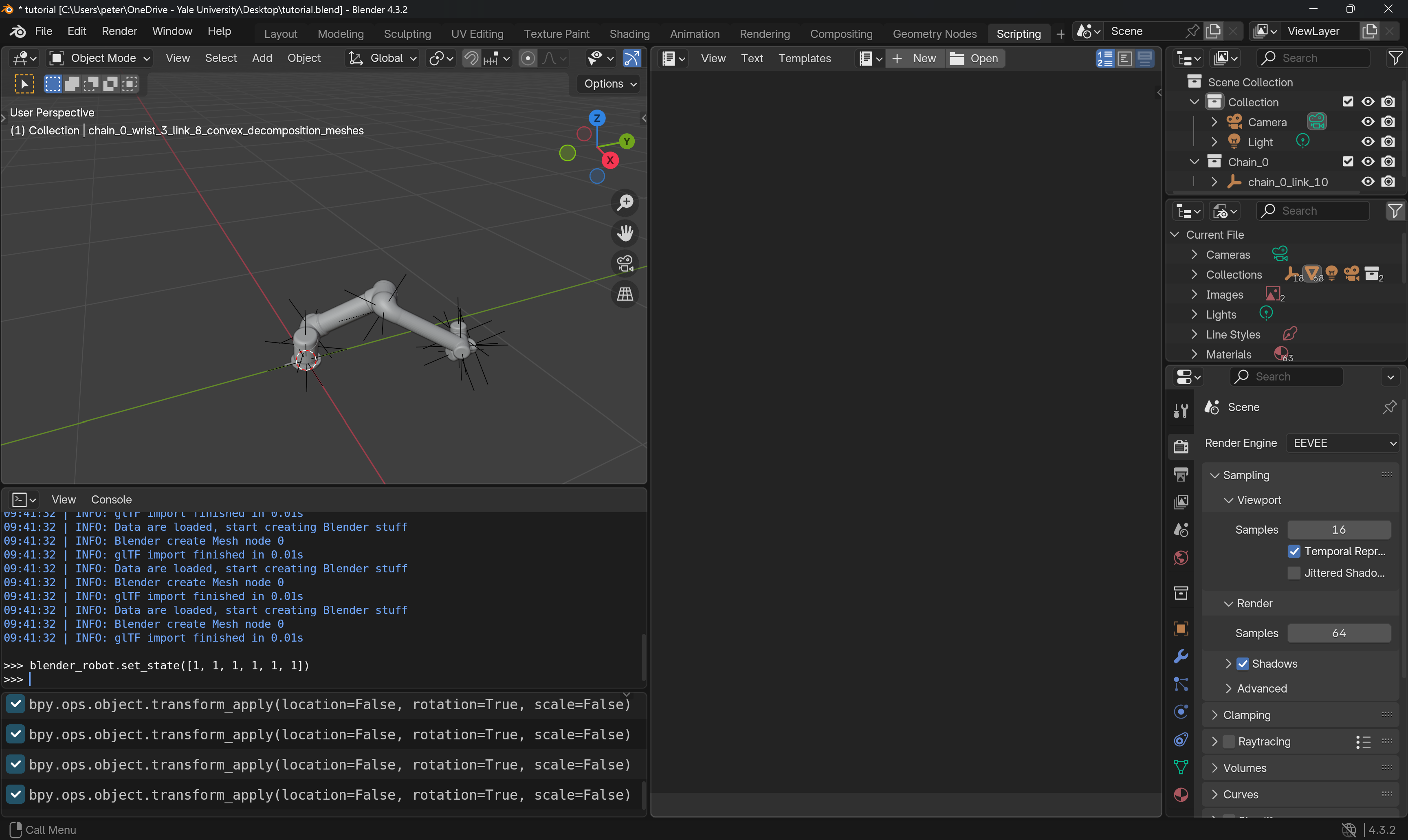}
    \caption{Example of a UR5 robot configured in Blender, where each joint angle has been set to 1 radian using a scripting call.}
    \label{fig:robot_config}
\end{figure}

Figure~\ref{fig:robot_config} illustrates this process for the UR5 robot, with all joints set to a uniform configuration of one radian. Beyond manually specifying states, the library also provides functions for forward and inverse kinematics, allowing users to compute the necessary joint angles for achieving target poses. This flexibility makes it straightforward to explore different robot configurations, whether for visualization, demonstration, or integration into larger animation sequences.

\subsection{Robot Visualization}

In addition to importing and configuring robots, the library provides flexible options for controlling how each robot and its components are visualized within Blender. Users can toggle the display of meshes, adjust transparency, and modify colors either globally or per-link. These tools make it straightforward to create clear, publication-quality images that highlight specific aspects of a robot’s structure.

\subsubsection{Plain Meshes}  
Each robot link can be represented using its raw mesh files (e.g., \texttt{.obj} or \texttt{.glb}). By default, these meshes are displayed, and users can further control their appearance through transparency and color settings. Transparency values range from 0 (fully transparent) to 1 (fully opaque), while colors are defined as RGBA values:

\begin{figure}[H]
    \centering
    \begin{lstlisting}[
        language=Python,
        frame=lines,
        basicstyle=\ttfamily\fontsize{7pt}{8pt}\selectfont,
        breaklines=true,
        label={lst:plain_mesh_style}
    ]
blender_robot.set_all_links_plain_mesh_alpha(alpha)
blender_robot.set_link_plain_mesh_alpha(
    link_index, alpha
)

blender_robot.set_all_links_plain_mesh_color([R, G, B, A])
blender_robot.set_link_plain_mesh_color(
    link_index, [R, G, B, A]
)\end{lstlisting}
    \caption{Commands for adjusting link transparency and color.}
    \label{fig:plain_mesh_style_code}
\end{figure}

\begin{figure}[t!]
    \centering
    \includegraphics[width=\columnwidth]{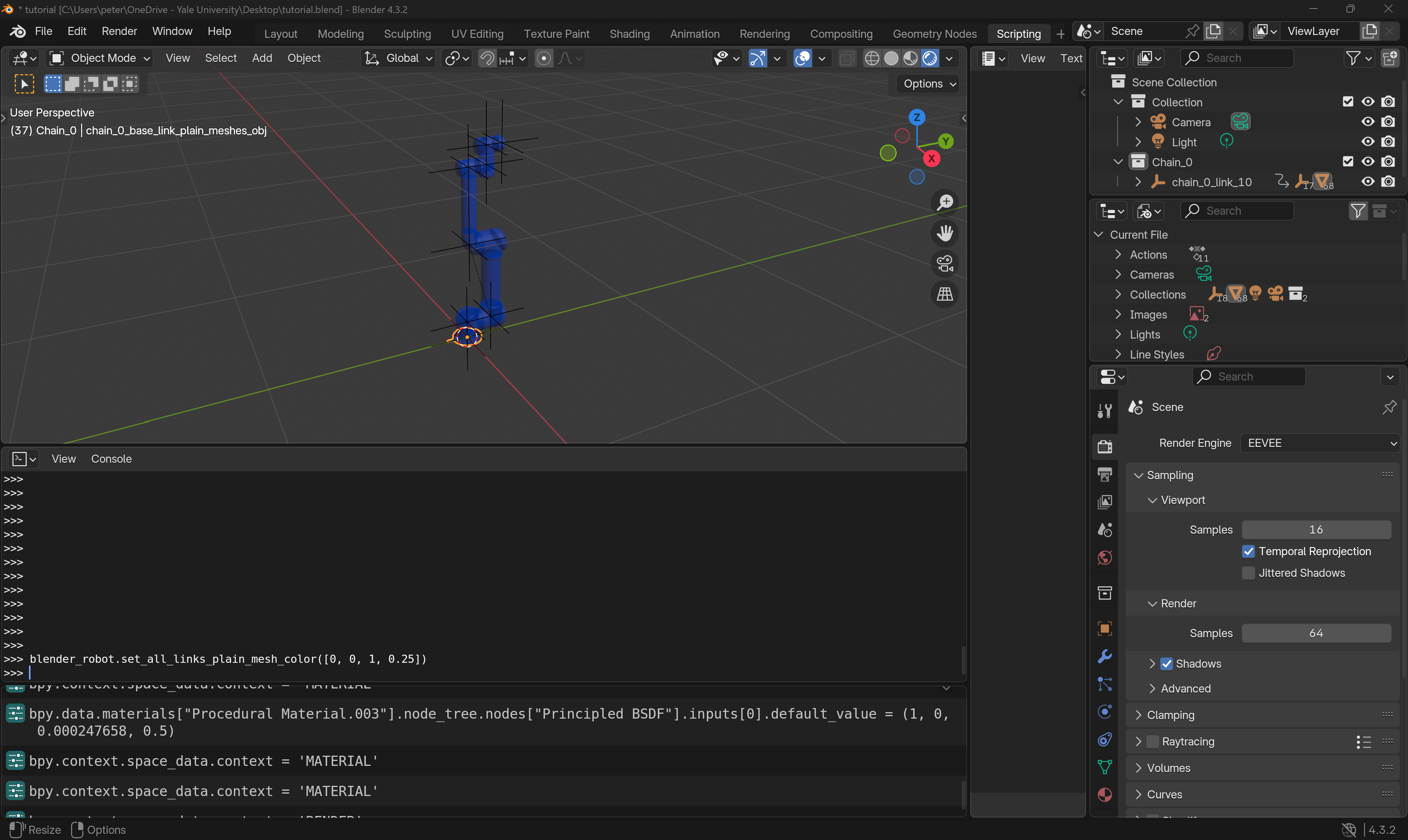}
    \caption{UR5 robot visualized with blue coloring and 0.25 transparency applied to all links.}
    \label{fig:link_color_transparency}
\end{figure}

Figure~\ref{fig:link_color_transparency} illustrates an example where the UR5 robot is displayed with blue coloring and partial transparency.  

\subsubsection{Convex Hulls}  
For simplified approximations of link geometry, the library supports convex hull visualization. Convex hulls can be toggled on or off and styled in the same way as meshes:

\begin{figure}[H]
    \centering
    \begin{lstlisting}[
        language=Python,
        label={lst:convex_style}
    ]
blender_robot.set_convex_hull_meshes_visibility(visible=True)

blender_robot.set_all_links_convex_hull_mesh_alpha(alpha)
blender_robot.set_link_convex_hull_mesh_alpha(
    link_index, alpha
)

blender_robot.set_all_links_convex_hull_mesh_color([R, G, B, A])
blender_robot.set_link_convex_hull_mesh_color(
    link_index, [R, G, B, A]
)\end{lstlisting}
    \vspace{-2pt}
    \caption{Commands for managing convex hull mesh visibility, alpha, and color.}
    \label{fig:convex_style_code}
\end{figure}

\begin{figure}[t!]
    \centering
    \includegraphics[width=\columnwidth]{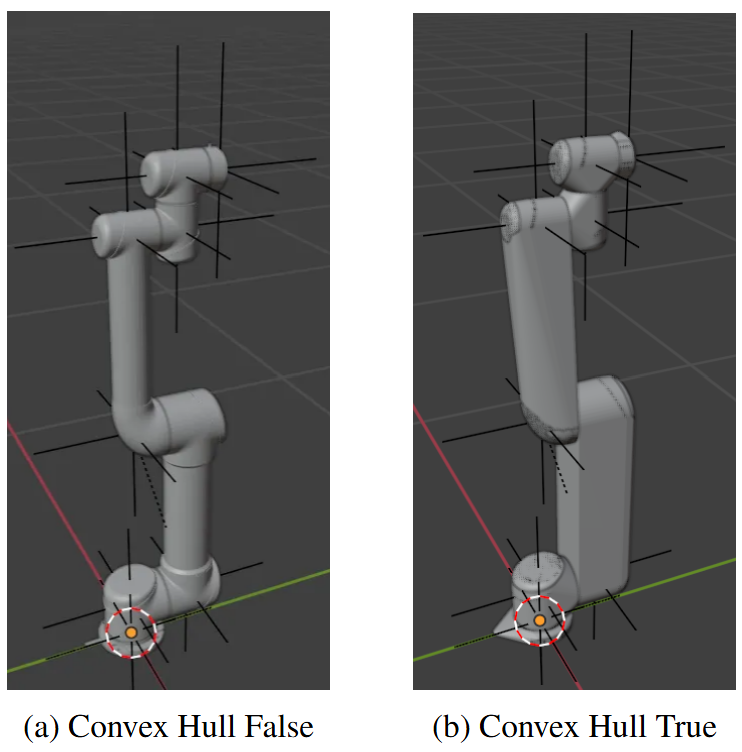}
    \caption{Convex hull visualization of the UR5 robot in Blender.}
    \label{fig:convex_hull}
    \vspace{-10pt}
\end{figure}

\subsubsection{Convex Decompositions}  
For higher-fidelity approximations, convex decompositions divide each link into multiple convex parts. As with meshes and convex hulls, decompositions can be toggled and styled with transparency and color:

\begin{figure}[H]
    \centering
    \begin{lstlisting}[
        language=Python,
        label={lst:decomp_style}
    ]
blender_robot.set_convex_decomposition_meshes_visibility(is_visible)

blender_robot.set_all_links_convex_decomposition_mesh_alpha(alpha_val)
blender_robot.set_link_convex_decomposition_mesh_alpha(
    idx, alpha_val
)

blender_robot.set_all_links_convex_decomposition_mesh_color([R, G, B, A])
blender_robot.set_link_convex_decomposition_mesh_color(
    idx, [R, G, B, A]
)\end{lstlisting}
    \vspace{-2pt}
    \caption{Commands for managing convex decomposition mesh visibility, alpha, and color.}
    \label{fig:decomp_style_code}
\end{figure}

\begin{figure}[t!]
    \centering
    \includegraphics[width=\columnwidth]{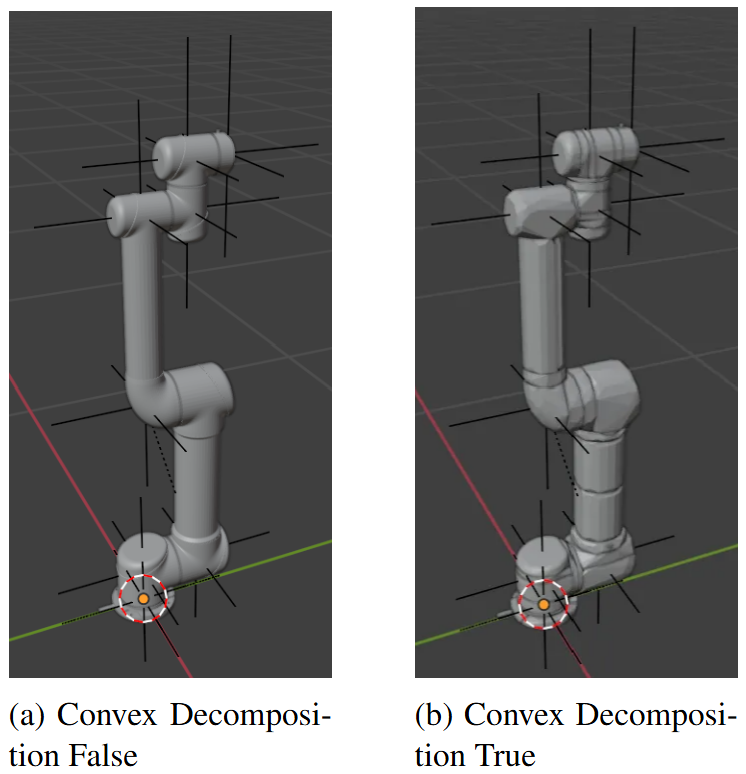}
    \caption{Convex decomposition visualization of the UR5 robot in Blender.}
    \label{fig:decomp}
    \vspace{-5pt}
\end{figure}

\subsection{Animation}

Animations in Blender are defined by \emph{keyframes}, which capture the state of an object at specific frames. Blender automatically interpolates between keyframes, generating smooth transitions. Our library extends this functionality to robots: users can define robot states at particular frames, and Blender will animate smooth, kinematically consistent motion between them. This enables dynamic demonstrations with minimal scripting effort.

\subsubsection{Configuration Keyframes}  
Robot joint configurations can be keyframed so that Blender interpolates poses over time. The core function is:

\begin{figure}[H]
    \centering
    \begin{lstlisting}[
        language=Python,
        label={lst:config_keyframe}
    ]
blender_robot.keyframe_state(frame_number)\end{lstlisting}
    \vspace{-2pt}
    \caption{Commands for keyframing the robot state in the Blender timeline.}
    \label{fig:config_keyframe_code}
\end{figure}

A typical workflow sets states at different frames and records them:

\begin{figure}[H]
    \centering
    \begin{lstlisting}[
        language=Python,
        label={lst:config_keyframe_ex}
    ]
blender_robot.set_state([0, 0, 0, 0, 0, 0])
blender_robot.keyframe_state(1)

blender_robot.set_state([1, 1, 1, 1, 1, 1])
blender_robot.keyframe_state(50)\end{lstlisting}
    \vspace{-2pt}
    \caption{Example of keyframing multiple robot states across the timeline.}
    \label{fig:config_keyframe_ex_code}
    \vspace{-10pt}
\end{figure}

Here, the robot starts at an all-zero configuration at frame 1 and moves to all-ones at frame 50. When played in Blender’s timeline, the intermediate frames are automatically interpolated.

For longer trajectories, users can keyframe an entire sequence at once:

\begin{figure}[H]
    \centering
    \begin{lstlisting}[
        language=Python,
        label={lst:config_keyframe_traj}
    ]
blender_robot.keyframe_discrete_trajectory(
    states_list # List[List[float]]
)\end{lstlisting}
    \vspace{-2pt}
    \caption{Commands for keyframing a discrete trajectory of states.}
    \label{fig:config_keyframe_traj_code}
\end{figure}

This function takes a list of states and inserts them sequentially into the timeline, avoiding the need to set each frame individually.

\subsubsection{Visual Attributes}  
In addition to joint configurations, keyframes can also be applied to visual properties such as color, transparency, or mesh visibility. By specifying different values at different frames, Blender automatically interpolates the changes — for example, smoothly fading a robot from opaque to transparent, or transitioning link colors from red to blue. These features provide further flexibility when creating animations for communication and demonstration.

\subsection{3D Shape Primitives}

In addition to robot models, the library provides scripting functions for generating auxiliary 3D primitives such as lines and cubes. While these shapes are simple, they serve as examples of a broader set of tools that allow users to annotate scenes, highlight geometric relationships, and build up higher-level visual elements. For instance, lines can represent arrows to indicate forces, torques, or directions of motion, while cubes can serve as bounding boxes for obstacles, workspace volumes, or regions of interest. Because primitives are tied to specific frames in the Blender timeline, their visibility can be controlled dynamically, enabling clear and pedagogical step-by-step visual explanations.

\subsubsection{Lines}  
Line objects are managed through a \texttt{BlenderLineSet}, which initializes a collection of lines and defines their properties at specified frames:

\begin{figure}[H]
    \centering
    \begin{lstlisting}[
        language=Python,
        label={lst:line_funcs}
    ]
lineset = BlenderLineSet(num_lines)
lineset.set_line_at_frame(
    start_pos, end_pos, frame,
    radius=0.01, color=[1, 0, 0, 1]
)\end{lstlisting}
    \vspace{-2pt}
    \caption{Functions for defining and keyframing individual lines in a set.}
    \label{fig:line_funcs}
\end{figure}

The first function creates the line set, while the second specifies the endpoints, radius, color, and the frame where the line becomes visible. 

\begin{figure}[t!]
    \centering
    \includegraphics[width=\columnwidth]{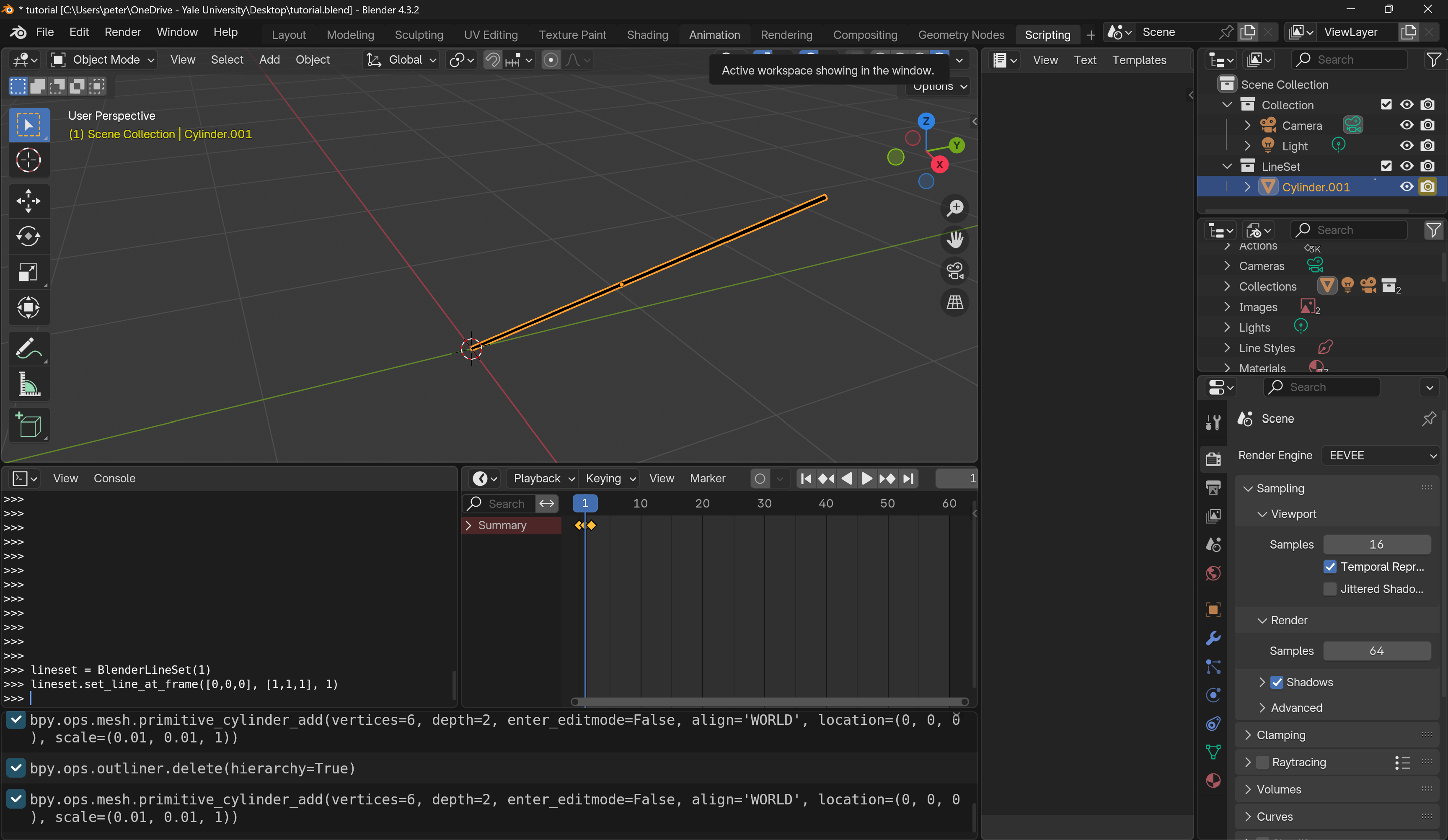}
    \caption{Example of a line defined in Blender via a script. Lines can be used to represent arrows or vector directions.}
    \label{fig:blender_line}
\end{figure}

This script creates a single line from the origin to the point \([1,1,1]\), visible at frame 1. Such a line can be interpreted as an arrow indicating motion or force. Figure~\ref{fig:blender_line} shows the resulting line rendered in Blender.

\subsubsection{Cubes}  
Cube objects are defined in a similar way using a \texttt{BlenderCubeSet}:

\begin{figure}[H]
    \centering
    \begin{lstlisting}[
        language=Python,
        label={lst:cube_funcs}
    ]
cubeset = BlenderCubeSet(num_cubes)
cubeset.set_cube_at_frame(
    center_pos, orientation_rpy,
    dimensions_lwh, frame,
    color=[1, 0, 0, 1], alpha=1.0
)\end{lstlisting}
    \vspace{-2pt}
    \caption{Functions for spawning and keyframing cube primitives within a set.}
    \label{fig:cube_funcs}
\end{figure}

Here the cube is defined by its center position, orientation (Euler angles in radians), dimensions, and optional color or transparency. 

\begin{figure}[t!]
    \centering
    \includegraphics[width=\columnwidth]{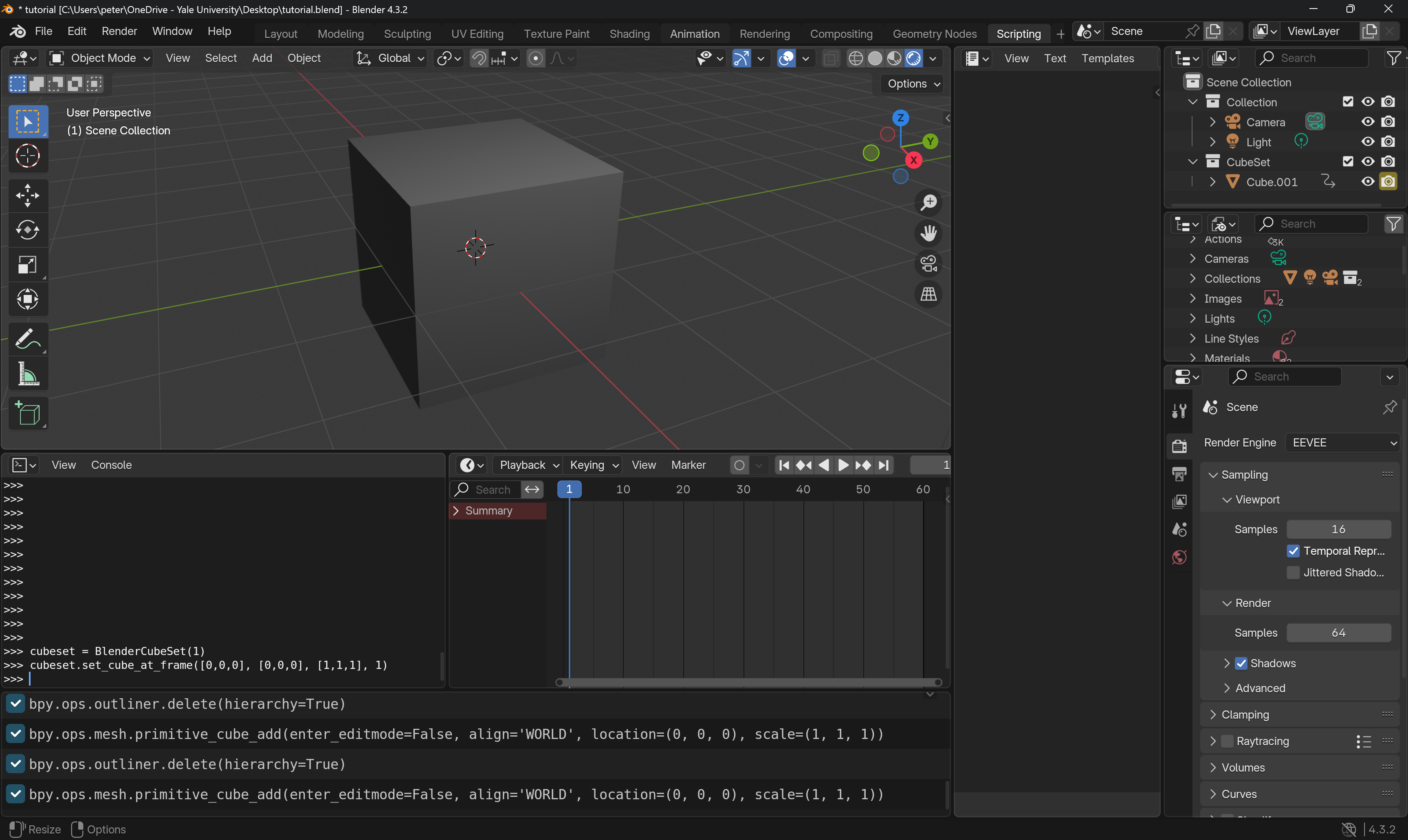}
    \caption{Example of a cube defined in Blender via a script. Cubes can serve as bounding boxes or workspace indicators.}
    \label{fig:blender_cube}
\end{figure}

This script spawns a unit cube at the origin with no rotation, visible at frame 1. Such cubes can be used to represent bounding boxes, approximate collision volumes, or highlight workspace regions. Figure~\ref{fig:blender_cube} shows the cube rendered in Blender.

\section{Evaluation and Results}
\label{sec:evaluation}

In this section, we demonstrate the capabilities of our system through a series of examples and benchmarks. The goal is to highlight how the library can be used to generate high-quality visualizations, streamline the process of configuring robots within Blender, and enable dynamic annotations that enrich educational and research figures.  In practice, versions of this library have been used to create figures and presentation materials associated with several previous papers\cite{liang2025ad, rakita2018autonomous, rakita2018relaxedik, rakita2018shared, rakita2019remote, rakita2019shared, rakita2019stampede, rakita2020analysis, rakita2020effects, rakita2021collisionik, rakita2021single, rakita2022proxima, rakita2025coherence}.  We direct interested readers to these papers to see how our library can be used in real-world academic settings.    



\subsection{Robot Motion Through Color Gradient}

\begin{figure}[t!]
    \centering
    \includegraphics[width=\columnwidth]{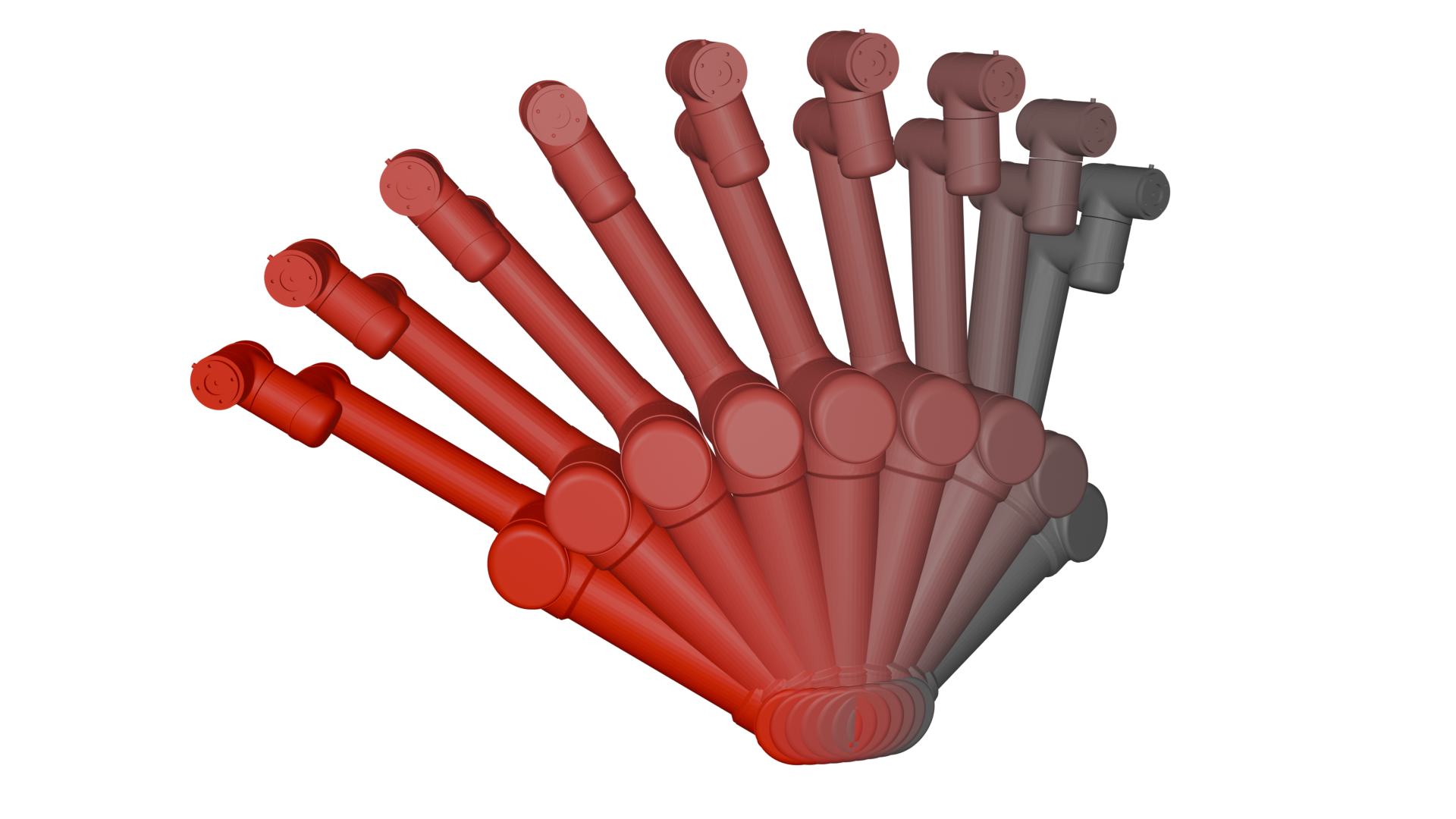}
    \caption{UR5 robot displaying motion through a color gradient.}
    \label{fig:ur5_motion}
\end{figure}

When publishing results, roboticists often need to convey robot motion through a single static image rather than an animation. Our system makes this process straightforward by providing scripting functions that allow motion to be illustrated via smooth color gradients. This approach produces clear, journal-quality figures that effectively communicate dynamic behavior in a compact visual form. Examples are shown in Figures~\ref{fig:teaser} and \ref{fig:ur5_motion}, which highlights how these gradients can be integrated into polished visualizations.

The example in Figure~\ref{fig:ur5_motion} was created with fewer than 100 lines of code. After importing the UR5 model, the user only needs to specify the start and end configurations, along with the number of intermediate copies to spawn. The following snippet demonstrates the core logic:

\begin{figure}[H]
    \centering
    \begin{lstlisting}[
        language=Python,
        label={lst:ur5_motion}
    ]
# robot configuration in joint space
start_state = np.array([0, -np.pi/3, np.pi/4, 0, 0, 0])
end_state   = np.array([-np.pi/4, np.pi/3, 0, 0, 0, 0])

# color gradient (RGBA)
start_color = np.array([0.2, 0.2, 0.2, 1])
end_color   = np.array([1, 0, 0, 1])

num_copies = 9
for i in range(num_copies):
    alpha = i / (num_copies - 1)
    state = start_state + (end_state - start_state) * alpha
    color = start_color + (end_color - start_color) * alpha
    robot = ChainBlender.spawn(c, r)
    robot.set_state(state)
    robot.set_all_links_plain_mesh_color(color)\end{lstlisting}
    \vspace{-2pt}
    \caption{Script for generating a motion sequence with a color gradient.}
    \label{fig:ur5_motion_code}
\end{figure}

This pattern generalizes to any robot model, enabling consistent and visually intuitive depictions of motion across platforms.

\subsection{Schematics for Robot Platforms}

\begin{figure}[t!]
    \centering
    \includegraphics[width=\columnwidth]{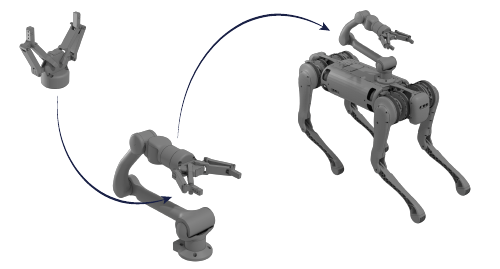}
    \caption{Schematic visualization of a Robotiq 140 Gripper, a Unitree Z1 Robot, and a Unitree B1 Robot coordinating together to form a single platform. All subcomponents of this image were generated and rendered using our library.}
    \label{fig:schematic}
\end{figure}

Clear schematics are a standard way of describing robot platforms, their components, and how they are assembled. With only a few lines of code, our system enables users to generate such schematics directly within Blender, blending geometric precision with professional-quality visualization. Figure~\ref{fig:schematic} shows an example schematic of an xArm7 with gripper and rail, created entirely through scripted functions.

\subsection{Visualizing Geometric Approximations}

Geometric approximations of robot links—such as convex hulls, convex decompositions, and bounding volumes—are foundational in collision checking, motion planning, and pedagogy. Clear visuals of these abstractions help readers understand why and how such approximations are used. Our system makes these views trivial to generate and style, enabling fast iteration on figures that balance accuracy and clarity.

\begin{figure}[t!]
    \centering
    \includegraphics[width=\columnwidth]{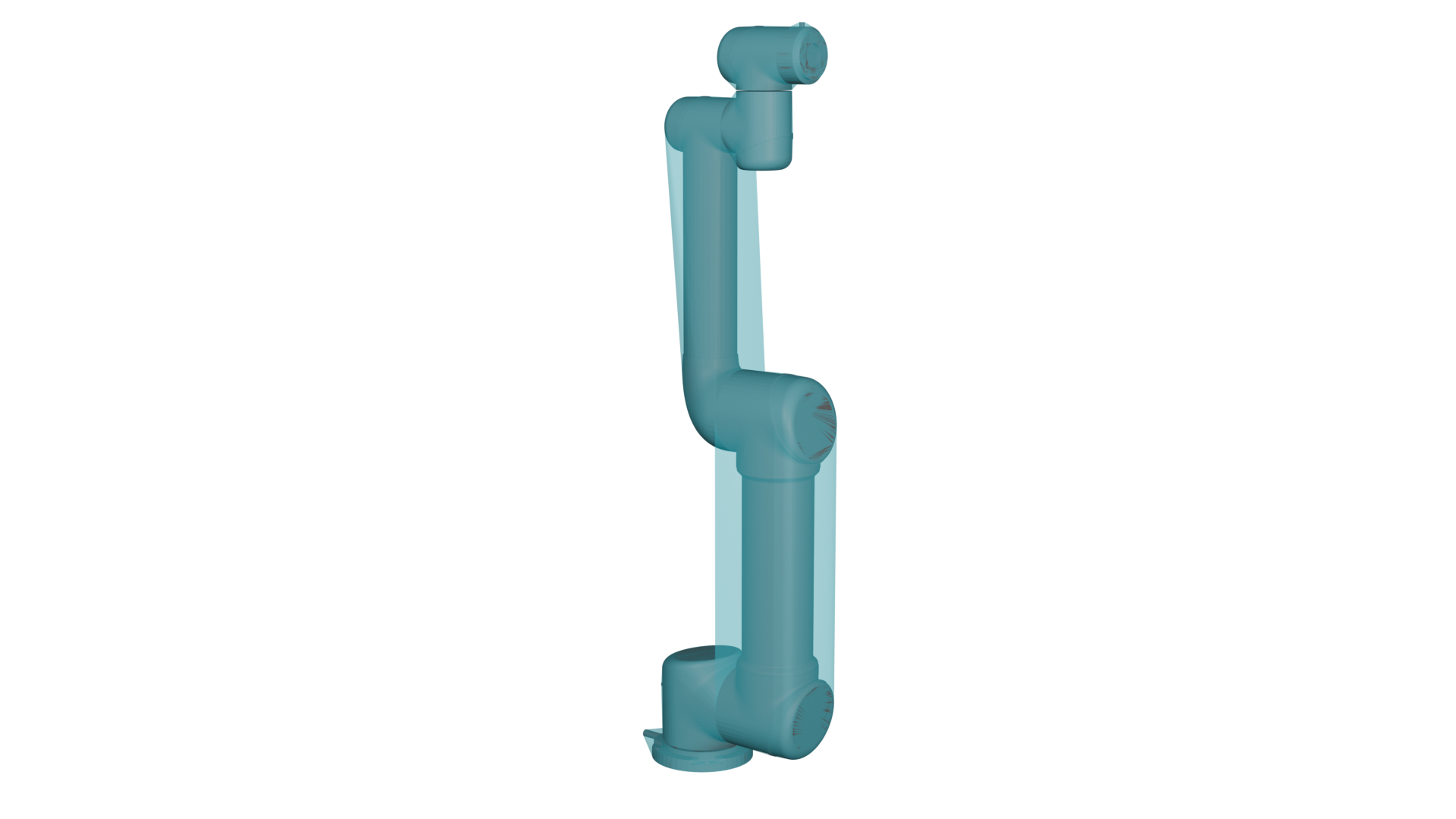}
    \caption{UR5 robot with per-link convex hulls visualized in Blender.}
    \label{fig:ur5_convex}
    \vspace{-5pt}
\end{figure}

Figure~\ref{fig:ur5_convex} shows a UR5 rendered with per-link convex hulls. The look (color, opacity) and scope (global vs.\ per-link) can be adjusted with a few lines of code, making it straightforward to tailor the visualization to the narrative of a paper or lecture.

\begin{figure}[H]
    \centering
    \begin{lstlisting}[
        language=Python,
        label={lst:ur5_convex}
    ]
# Toggle convex hulls and set a consistent style
blender_robot.set_convex_hull_meshes_visibility(True)

blender_robot.set_all_links_convex_hull_mesh_color(
    [0.02, 0.81, 0.98, 1.0] # RGBA
)

blender_robot.set_all_links_convex_hull_mesh_alpha(0.50)\end{lstlisting}
    \vspace{-2pt}
    \caption{Scripting interface for toggling and styling UR5 convex hulls.}
    \label{fig:ur5_convex_code}
\end{figure}

This workflow generalizes across platforms: with the same pattern, users can swap to convex decompositions or bounding boxes, adjust transparency to reveal internal structure, and apply per-link styling to highlight areas of interest. In practice, producing a polished approximation view requires only a short script, yielding reproducible figures suitable for both evaluation and instruction.

\section{Discussion}
\label{sec:discussion}

This work aims to lower the barrier to producing clear, publication-quality visuals in robotics by embedding robotics-aware scripting directly within Blender. Rather than competing with simulators or game engines that prioritize physical fidelity or real-time interaction, our library is intentionally designed around three pillars: high-quality offline rendering, a concise Python interface, and compatibility with standardized robot description formats (e.g., URDF).

The examples shown in \S\ref{sec:evaluation} demonstrate how these design choices translate into practical benefits. Motion-through-color gradients compress temporal behavior into a single image; schematic illustrations overlay lightweight primitives atop robot geometry to communicate mechanical structure or subsystem roles; and approximating geometries such as convex hulls and decompositions can be rendered and styled for pedagogy, collision reasoning, or planning visualization. Because installation targets Blender’s bundled Python using a single pip command, the setup cost is minimal, making it practical for both classroom instruction and collaborative multi-lab projects.

Beyond convenience, a scripting-first workflow significantly improves reproducibility. All figures and videos are generated from code that specifies the exact scene layout, camera configuration, materials, and rendering parameters, enabling version control and seamless regeneration as models or datasets evolve. The same script can produce both stills and animations, ensuring consistency in framing, lighting, and materials across formats. For instructional use, frame-accurate annotations, such as arrows, coordinate axes, and bounding boxes, can be programmatically placed on robot geometry to support step-by-step explanations that emphasize spatial reasoning over software tooling.

The design philosophy emphasizes minimalism and composability. The core API remains small—focusing on import, configuration, keyframing, and primitive generation—so users can build their own extensions (e.g., custom camera paths, lighting presets, annotation utilities) without modifying the library itself. Open-source distribution further supports community contributions, shared templates, and streamlined bug reporting.

\subsection{Limitations}

Our work is subject to specific limitations that highlight opportunities for future development. Primarily, this library functions strictly as a visualization tool, not a physics engine; it does not model dynamics, contact interactions, or closed-loop control. Consequently, interpolation between keyframes is purely kinematic unless driven by external trajectory data. The system also relies on well-formed robot description files, meaning malformed or incomplete URDFs may require manual correction prior to import. regarding performance, high-fidelity scenes, characterized by high-polygon meshes or dense convex decompositions, can increase memory overhead and render times, reflecting the constraints of Blender's offline renderer. Furthermore, while the library streamlines scene construction, users still require a working knowledge of Blender fundamentals, such as lighting and camera composition, to achieve high-quality results. Finally, installation via pip may vary across operating systems and Blender versions. Ultimately, we design this tool to complement, rather than replace, simulators; it is most effective when paired with external systems for planning and data generation.






\bibliographystyle{plainnat}

\bibliography{refs}


\end{document}